Research Article

Running Title: Bayesian Causal AI to Evaluate Air Pollution Regulations in London

# AI for Social Good: An Investigation of the Causal Relationship Between Environmental Regulations and Their Effects on Air Pollution in London, UK

Author 1: Dr. Yang Han [1#]

Author 2: Prof. Jacqueline CK Lam [1#*]

Author 3: Prof. Victor OK Li [1#*]

Author 4: Mr. Yiu-Wai Man [1#]

[1]Department of Electrical and Electronic Engineering, The University of Hong Kong, Hong Kong

#Equal contributions.

*Corresponding authors: jcklam@eee.hku.hk; vli@eee.hku.hk

## Abstract

Air pollution regulation is central to urban public health governance, but estimating its effects is difficult because policies are implemented non-randomly and pollution trajectories are shaped by meteorology, socioeconomic change, temporal trends, and overlapping interventions. This study develops an uncertainty-aware Bayesian deep learning framework to estimate the aggregate effect of air pollution regulations on $PM_{2.5}$ concentrations in London from 2010 to 2020. The framework integrates daily $PM_{2.5}$ observations from Inner London monitoring stations, meteorological covariates, annual socioeconomic indicators, month-of-year and day-of-week indicators, and daily regulation status data for 32 policy measures. A Bayesian LSTM captures temporal dependencies in environmental and socioeconomic covariates, Bayesian embedding layers represent temporal and regulation status inputs, and a regulation status prediction branch supports propensity score-based adjustment for non-random policy implementation. Regulatory effects are estimated by comparing observed $PM_{2.5}$ concentrations with counterfactual predictions under a hypothetical no-regulation scenario, with uncertainty summarized across repeated Bayesian training runs and bootstrap resampling. Results show that London's regulations were associated with an average $PM_{2.5}$ reduction of 1.88 $\mu g/m^3$, a relative reduction of 12.35%, with a 95% confidence interval of 1.64-2.12 $\mu g/m^3$. Estimated effects were limited before 2013, became clearer from 2013 to 2017, and were strongest in 2018 and 2019. The findings suggest that sustained and cumulative regulatory interventions contributed to measurable improvements in London's air quality. This study demonstrates how uncertainty-aware causal AI can support environmental accountability, public health protection, and evidence-based governance for environmental decision-making.

Policy Significance Statement

This study provides evidence that London's air pollution regulations were associated with measurable reductions in $PM_{2.5}$ concentrations, especially after 2013 and more strongly in 2018-2019. By estimating counterfactual pollution levels under a hypothetical no-regulation scenario, this study offers policymakers a clearer basis for assessing whether regulatory interventions contributed to air quality improvement. The findings support sustained, cumulative, and evidence-based air pollution governance rather than short-term or isolated policy action. Methodologically, the uncertainty-aware Bayesian deep learning framework helps quantify both estimated regulatory benefits and uncertainty, making it useful for transparent policy evaluation. Overall, this study shows how causal AI can support regulatory accountability, public health protection, and data-driven urban environmental decision-making.

## 1. Introduction

Air pollution is a major public health and governance challenge in large cities. Rapid socioeconomic development, globalization, urbanization, and population growth have intensified environmental pressures, while transport activity and economic production continue to generate harmful emissions. Among different air pollutants, fine particulate matter, particularly $PM_{2.5}$, is important because it can penetrate deeply into the respiratory system and is associated with serious health risks (Guyatt et al. 2024; Vanoli et al. 2024). In London, the public health consequences of air pollution have become increasingly visible. The death of Ella Adoo-Kissi-Debrah, which was legally linked to excessive air pollution exposure in southeast London, highlighted the urgent need for effective air quality governance and policy accountability (BBC 2021). Over the past decade, London has implemented multiple air pollution control measures, including Low Emission Zone (LEZ) and Ultra Low Emission Zone (ULEZ) policies. London therefore provides an important case for evaluating whether urban air pollution regulations are associated with measurable reductions in $PM_{2.5}$ concentration.

Estimating regulatory effects is difficult because environmental policies are not implemented randomly. Air pollution regulations are usually introduced in response to existing pollution levels, public health risks, transport conditions, socioeconomic pressures, and political priorities, meaning that regulation exposure is often shaped by prior conditions rather than randomized assignment (Mork et al. 2024). As a result, observed changes in $PM_{2.5}$ concentration after policy implementation cannot be interpreted as regulatory effects without accounting for confounding bias. In air pollution policy evaluation, key confounders include meteorology, economic activity, transport intensity, seasonal trends, and long-term urban change, which may affect both regulatory decisions and pollution concentration (Dai et al. 2024; Tec et al. 2023).

Randomized controlled trials can provide strong causal evidence, but they are often costly, impractical, or infeasible in large-scale public policy settings (Dahabreh and Bibbins-Domingo 2024). As a result, air pollution policy evaluation commonly depends on observational data, where regulation exposure is not randomly assigned. Traditional statistical approaches often estimate associations between treatments and outcomes, but association alone is insufficient for regulatory effect estimation when policy implementation is shaped by factors such as pollution levels, public health concerns, transport conditions, meteorology, and socioeconomic change (Brewer et al. 2023). Causal inference provides a stronger framework for observational policy evaluation by estimating policy effects through counterfactual comparisons between treatment and non-treatment conditions (Leist et al. 2022). The Rubin Causal Model formalizes this logic through potential outcomes, where causal effects are defined by comparing the outcome observed under treatment with the counterfactual outcome that would have occurred under non-treatment (Rubin 2005; 2025). Building on this counterfactual estimation framework, propensity score matching and related adjustment methods seek to reduce confounding bias by improving comparability between treated and untreated observations (Abadie and Imbens 2016; Baccini et al. 2017; Sant'Anna et al. 2022).

Other observational causal inference methods, including instrumental variables, difference-in-differences, and regression discontinuity, are also widely used in policy evaluation, but each faces important limitations in air pollution regulation settings. Instrumental variable methods require variables that affect treatment assignment but influence the outcome only through treatment (Baiocchi et al. 2014; Schwartz et al. 2016), yet valid instruments are difficult to identify because many candidate instruments may also directly affect air quality. Difference-in-differences methods compare treated and untreated groups before and after policy implementation (Goodman-Bacon 2021; Goodman-Bacon and Marcus 2020; Liu et al. 2022), but they require credible control groups and parallel pre-treatment trends, which may be difficult to establish when regulations overlap across time, geography, and policy domains. Regression discontinuity designs estimate treatment effects near a treatment-assignment threshold (Tan and Mao 2021), yet many air pollution regulations do not have a clear cut-off rule, and selecting a threshold may itself introduce bias. These limitations mean that confounding remains a major challenge in large-scale observational environmental datasets, even when causal inference methods are used (Tec et al. 2023).

Synthetic control and related panel data counterfactual methods provide another important strand of policy evaluation. These approaches estimate counterfactual outcomes by leveraging information from untreated units observed over time. Synthetic control methods construct a weighted combination of untreated units to closely reproduce the treated unit's pre-intervention trajectory, making them particularly useful when a policy is implemented in one or a small number of treated areas. Building on this logic, synthetic difference-in-differences combines the unit-weighting framework of synthetic control with the time-adjustment structure of difference-in-differences (Arkhangelsky et al. 2021). Augmented synthetic control further improves performance when pre-treatment fit is imperfect by incorporating outcome modeling into the estimation procedure (Ben-Michael et al. 2021). More broadly, these approaches connect to a wider class of panel data imputation methods that estimate missing untreated potential outcomes using the observed outcome matrix. Matrix completion methods provide one such flexible approach, exploiting latent interactive structure across units and time to estimate untreated potential outcomes in panel data (Athey et al. 2021).

These causal inference methods have been increasingly applied in air pollution policy evaluation, particularly in studies of low-emission policies. Evidence from London suggests that LEZ and ULEZ policies reduced traffic-related pollutants, especially $NO_2$, but produced more limited or harder-to-detect effects for $PM_{2.5}$ because of its diffuse traffic, regional, and non-traffic sources (Mudway et al. 2019; Prieto-Rodriguez et al. 2022; Tong et al. 2025). Related European evidence also indicates that low-emission zones can generate health benefits, including reductions in cardiovascular morbidity and children's medication use (Klauber et al. 2024; Margaryan 2021). Together, these studies show that low-emission policies can improve traffic-related air quality and health, while highlighting the difficulty of identifying $PM_{2.5}$ impacts and cumulative regulatory effects.

Recent advances in machine learning and deep learning create new opportunities for policy evaluation using complex data. Machine learning-based causal inference methods can model nonlinear relationships and high-dimensional covariates more flexibly than many conventional statistical models (Feuerriegel et al. 2024; Moccia et al. 2024). Double/debiased machine learning separates flexible nuisance-function estimation (e.g., outcome regressions and propensity scores) from the target causal parameter and can support high-dimensional adjustment when the underlying causal design is credible (Chernozhukov et al. 2018; Chernozhukov et al. 2022; Fuhr et al. 2024). Causal forests combine random forests with the potential outcome framework to estimate heterogeneous treatment effects nonparametrically (Wager and Athey 2018), and related work has extended or adapted this method to environmental, spatial, and policy evaluation settings (Credit and Lehnert 2024; Miller 2020). Other studies have used recurrent neural networks, deep forecasting algorithms, and ensemble learning to predict counterfactual outcomes in panel data (Athey et al. 2019; Goldin et al. 2022; Poulos and Zeng 2021). These approaches demonstrate the potential of data science for policy evaluation, but their application to air pollution regulation remains challenging because air quality is shaped by temporal dependence, meteorological variation, socioeconomic activity, spatial transport, and multiple overlapping regulations (Dai et al. 2026; Heffernan et al. 2025; Qiu et al. 2022).

Bayesian deep learning is particularly relevant for this problem because it combines flexible prediction with uncertainty representation. Instead of learning fixed numerical weight parameters, Bayesian neural networks learn probability distributions over model parameters (Blundell et al. 2015). This is valuable for noisy, high-dimensional, and imbalanced real-world environmental datasets, where overfitting and model instability can otherwise lead to misleading estimates (Magris and Iosifidis 2023).

Bayesian recurrent neural networks extend this advantage to sequential settings, including models with long short-term memory (LSTM) units that capture temporal dependencies while propagating uncertainty through time (Fortunato et al. 2017). These properties are especially important for air pollution policy evaluation, where observed pollution trajectories reflect not only regulatory interventions but also time-varying meteorological and socioeconomic conditions that may confound estimated policy effects (Liang et al. 2017; Ren and Matsumoto 2020). A Bayesian temporal modeling framework is therefore well suited to estimating counterfactual pollution trajectories while representing uncertainty in complex, dynamic, and confounded environmental systems.

The most directly related work is Han et al. (2021a), which developed a Bayesian LSTM framework to evaluate the effects of air pollution control regulations in Beijing, China. This prior work integrated proxy data, including aerosol optical depth, meteorology, and socioeconomic variables, and used propensity score estimation to reduce confounding in regulatory effect analysis. It predicted counterfactual $PM_{2.5}$ concentrations under a hypothetical no-regulation scenario and estimated the average reduction in $PM_{2.5}$ attributable to air pollution control regulations, demonstrating the potential of Bayesian recurrent modeling for data-driven regulatory intervention analysis. However, because this framework was developed and tested in the Beijing regulatory context, its transferability to other urban governance systems and policy regimes remains an open question.

Until now, three gaps remain in data-driven air pollution policy evaluation. First, the external validity of Bayesian recurrent counterfactual frameworks for air pollution policy evaluation remains under-examined. Han et al. (2021a) demonstrated the feasibility of a Bayesian LSTM approach in Beijing, but it remains unclear whether such a framework can be transferred to a different urban governance context with different policy instruments, regulatory geography, transport governance, data structures, and intervention timelines. London differs from Beijing in regulatory structure, policy geography, transport-oriented governance, data availability, and intervention design, including low-emission and ultra-low-emission zones. Examining this transferability is important because data-driven policy evaluation methods must be adaptable across different institutional and urban settings. Second, existing London evidence has focused mainly on specific low-emission or ultra-low-emission transport policies, often emphasizing traffic-related pollutants such as $NO_2$ (Mudway et al. 2019; Prieto-Rodriguez et al. 2022; Tong et al. 2025). The aggregate effect of London's broader regulatory package on $PM_{2.5}$ remains less directly examined. Third, many machine learning applications in air pollution research remain prediction-oriented rather than policy-evaluation-oriented (Houdou et al. 2024). Accurate prediction is useful, but environmental governance requires evidence about whether regulations changed pollution trajectories, what pollution levels might have been under alternative policy scenarios, and how uncertain those estimates are. From an AI for Social Good perspective, this distinction is critical: AI systems designed for public benefit should not only produce accurate forecasts, but should also generate interpretable, uncertainty-aware evidence that supports policy accountability, public health protection, and transparent environmental decision-making.

To address these gaps, this study develops an uncertainty-aware Bayesian deep learning framework to estimate aggregate regulatory effects on $PM_{2.5}$ concentration in London using observational data. The study covers a ten-year period from 1 January 2010 to 31 January 2020. The model integrates daily air quality observations, meteorological covariates, socioeconomic indicators, month-of-year and day-of-week indicators, and regulation status data. A Bayesian LSTM component captures temporal dependencies in air pollution and proxy covariates, while Bayesian embedding layers represent month-of-year, day-of-week, and regulation status indicators. The model produces two outputs: predicted $PM_{2.5}$ concentration and predicted probability of regulation status. The regulation-status prediction branch supports propensity score-based adjustment for non-random regulation implementation. Regulatory effects are then estimated through counterfactual analysis by comparing observed $PM_{2.5}$ concentrations with counterfactual $PM_{2.5}$ concentrations under a hypothetical no-regulation scenario.

This study makes four contributions to data-driven policy evaluation and AI for Social Good. First, it extends the Bayesian LSTM regulatory intervention framework (Han et al. 2021a) from Beijing to London, thereby examining its applicability across distinct urban governance and regulatory contexts. Second, it constructs a London-specific, temporally aligned policy evaluation dataset linking $PM_{2.5}$ observations, meteorological conditions, socioeconomic indicators, month-of-year and day-of-week indicators, and regulation status information over a ten-year study period. Third, it estimates aggregate regulatory effects through counterfactual no-regulation prediction while using a regulation status prediction branch to support propensity score-based adjustment for non-random policy implementation. Fourth, it demonstrates how uncertainty-aware machine learning can serve socially beneficial policy goals by transforming complex urban data into evidence for environmental accountability, public health protection, and data-driven governance.

More broadly, this study shows how uncertainty-aware causal AI can support data-driven evidence-based environmental decision-making. Air pollution regulation is not only a technical policy problem, but also a public health, social welfare, and environmental justice issue. In cities where multiple regulations operate simultaneously, conventional evaluation designs may be difficult to apply, and policymakers may lack clear evidence about the aggregate consequences of regulatory action. By combining environmental, socioeconomic, and policy data with Bayesian uncertainty estimation and counterfactual prediction, the proposed framework provides interpretable evidence for policy accountability, environmental justice, and public health protection. In doing so, the study positions AI not merely as a forecasting tool, but as a decision-support approach for evaluating whether public interventions deliver measurable social benefits.

The rest of this paper is organized as follows. Section 2 describes the data collection procedure and the proposed Bayesian deep learning methodology. Section 3 reports the empirical results. Section 4 discusses the key findings, policy implications, and AI for Social Good significance. Section 5 concludes the study.

## 2. Data and Method

### 2.1. Data Collection

Data on air quality, meteorology, socioeconomic conditions, and air pollution regulation status were collected for the period from 1 January 2010 to 31 January 2020. The study period ends before the main COVID-19 period to exclude potential confounding effects from COVID-19-related restrictions on air quality. Table 1 summarizes the variables, spatial scales, temporal resolutions, and data sources.

**Table 1.** Data description

| Data | Variable(s) | Spatial | Temporal | Source |
|---|---|---|---|---|
| Air quality | $PM_{2.5}$ concentration | Station level | Daily | London Air (2022) |
| Meteorology | • Wind speed<br>• Wind direction<br>• Temperature<br>• Air pressure<br>• Relative humidity<br>• Visibility | City level | Daily | Meteostat (2022); NOAA (2022) |
| Socio-economic statistics | • Population density<br>• Number of licensed vehicles | City level | Yearly | Department for Transport (2020); Office for National Statistics (2020) |
| Air pollution regulation status | Enforcement periods of major air pollution control measures | City level | Daily | Defra (2015; 2021); The National Archives (2022) |
| **Note.** Relative humidity was not directly available from the selected data sources. Dew-point temperature and air temperature were used to estimate relative humidity. | | | | |

#### 2.1.1. Air Quality Data

$PM_{2.5}$ was selected as the outcome variable because fine particulate matter is strongly associated with adverse health effects, particularly respiratory and cardiovascular outcomes (Guyatt et al. 2024; Vanoli et al. 2024). Daily $PM_{2.5}$ concentration data, measured in μg/m$^3$, were obtained from the London Air Quality Network for the period from 1 January 2010 to 31 January 2020 (London Air 2022). This station-level data source provides finer spatial resolution than city-level aggregate data. Only measurements with reference-equivalent labels were retained to ensure data quality. To improve the relevance of the air quality outcome to air pollution control regulations, the analysis focused on 21 monitoring stations located in Inner London. This spatial restriction is appropriate because several major air pollution control measures, such as the Congestion Charge and the Ultra-Low Emission Zone, operated wholly or partly within Inner London during the study period. The station-level observations were aggregated into daily mean $PM_{2.5}$ concentrations across the selected stations during the study period (see Figure 1).

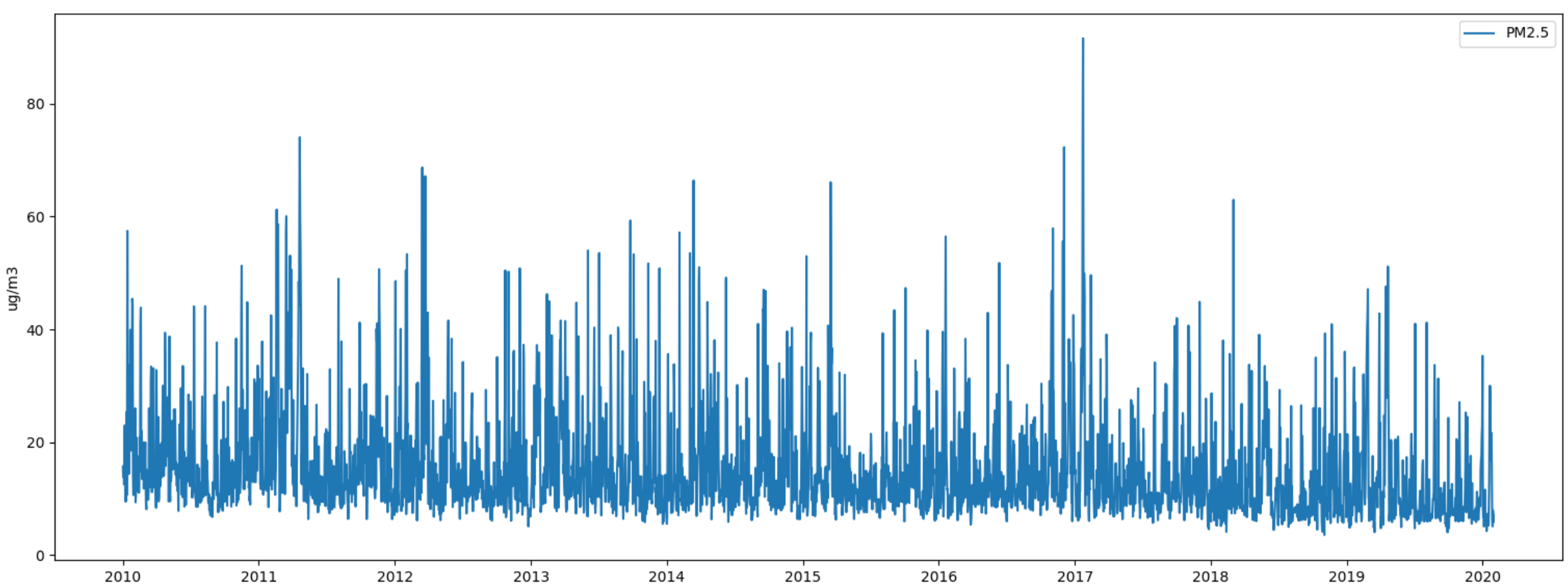

**Figure 1.** Aggregated daily mean $PM_{2.5}$ concentrations in μg/m$^3$ during the study period

### 2.1.2. Meteorology Data

Meteorological variables were included as proxy covariates because weather conditions can strongly influence air pollution concentration and dispersion. Prior studies have shown that meteorological data can improve probabilistic modeling of air pollution regulation effects (Liang et al. 2017). Daily city-level air temperature (°C), dew-point temperature (°C), visibility (miles), wind speed (knots) data from 1 January 2010 to 31 January 2020 were obtained from the Integrated Surface Database (ISD) of the U.S. National Oceanic and Atmospheric Administration (NOAA) (NOAA 2022). Daily wind direction (degree) and air pressure (hPa) data from 1 January 2010 to 31 January 2020 were obtained from an open weather and climate data vendor (Meteostat 2022). Because relative humidity was not directly available from credible sources, it was estimated from air temperature and dew-point temperature using Equations (1) and (2). Equation (1) computes saturation vapor pressure within the temperature range from -40°C to 50°C (Alduchov and Eskridge 1996). Relative humidity was then calculated by comparing saturation vapor pressure at dew-point temperature with saturation vapor pressure at air temperature. Figure 2 shows the daily meteorological variables used in the analysis.

$$e_w(\text{temp}) = 6.1094 \times e^{17.625t/(243.04+t)} \tag{1}$$

$$RH = 100 \times \frac{e_s(\text{temp}_{\text{dew point}})}{e_s(\text{temp})} \tag{2}$$

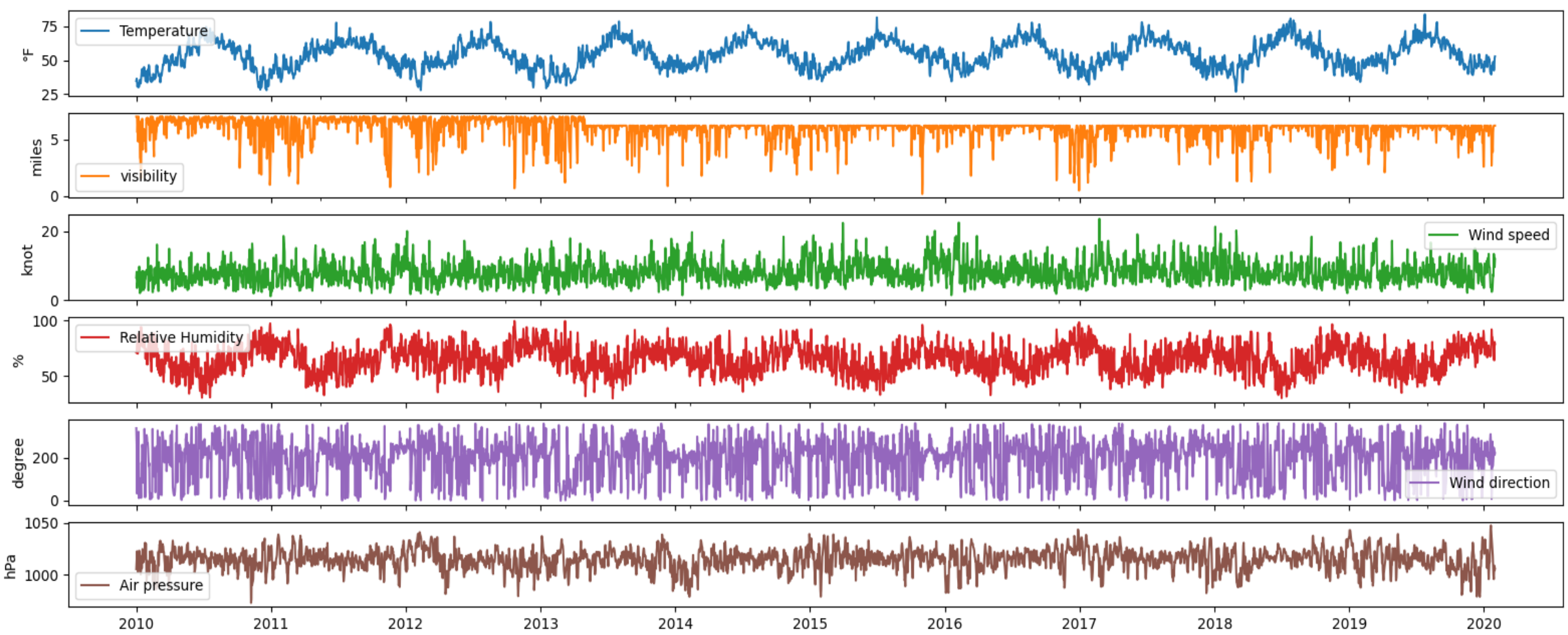

**Figure 2.** Daily meteorology data during the study period

### 2.1.3. Socio-economic Data

Socioeconomic variables were incorporated because previous studies have shown that demographic and transportation-related covariates can improve model performance when estimating the effect of air pollution regulations on air quality (Zheng et al. 2015). Annual socio-economic statistics from 1 January 2010 to 31 January 2020 were collected. Population data were obtained from the 2020 mid-year estimate by Office for National Statistics (Office for National Statistics 2020). Because several air pollution control measures were concentrated in Inner London, the population of Inner London boroughs was aggregated and divided by the Inner London land area of 319 $km^2$ to calculate population density. The number of licensed vehicles was obtained from the Department for Transport (Department for Transport 2020). All types of vehicles were included.

### 2.1.4. Regulation Status Data

The regulation dataset was constructed from annual reports, policy documents, and official government sources. Enforcement dates, amendment dates, revision dates, and end dates were obtained primarily from UK legislation records and official policy documents (Defra 2015; 2021; The National Archives 2022). Where a policy remained active beyond the study period, it was treated as active through 31 January 2020. Major air pollution control policies and regulatory measures implemented between 2010 and 2020 were identified and coded (see Figure 3). These measures include national regulations implementing European Union directives, such as the Air Quality Standards Regulations 2010, which implement the Ambient Air Quality Directive 2008/50/EC and the Fourth Daughter Directive 2004/107/EC. They also include London-specific policy measures, such as the Ultra-Low Emission Zone.

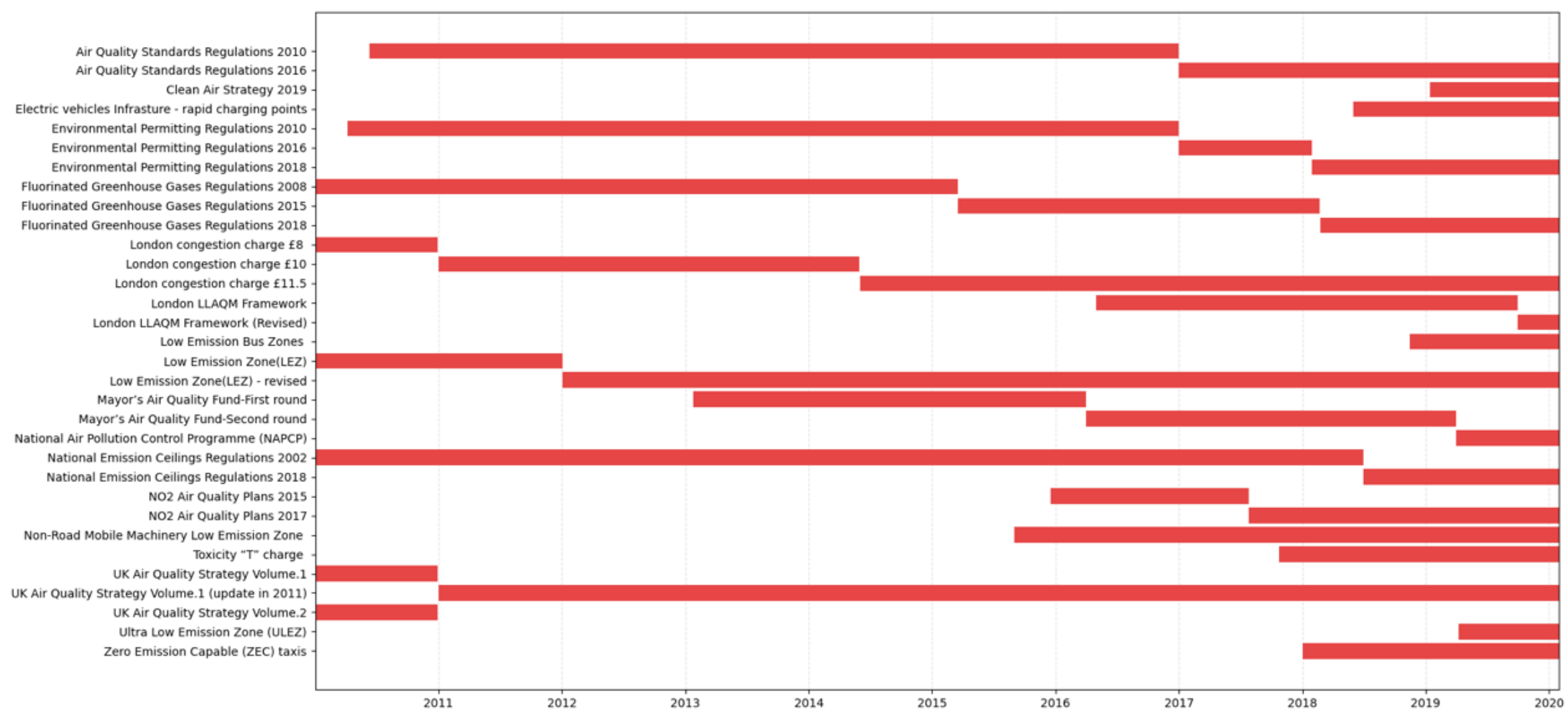

**Figure 3.** Major air pollution control regulations from 2010 to 2020, including national policies implemented under the EU directives and action plans in boroughs belonging to Inner London

## 2.2. Data Preprocessing

The regulation dataset contains 32 major air pollution control policies or policy revisions. Each policy record includes an enforcement start date and, where applicable, an end date. These records were transformed into a daily regulation status matrix covering all dates from 1 January 2010 to 31 January 2020. For each day $t$, regulation status is represented as a binary vector: $I_t = \{I_t^1, \dots, I_t^k\}$, where $I_t^k = 1$ indicates that regulation $k$ was enforced on day $t$, and $I_t^k = 0$ indicates that it was not enforced.

Air quality and meteorological observations were aggregated to the daily level by calculating daily mean values. Annual socioeconomic variables, including population density and number of licensed vehicles, were assigned to all daily observations within the corresponding year. The final analytical dataset contains daily $PM_{2.5}$ concentration, meteorological covariates, annual socioeconomic covariates, temporal indicators, and daily regulation status indicators.

The dataset was divided into training, validation, and test sets. During the full study period, 80% of the daily observations were assigned to the training set, 10% to the validation set, and 10% to the test set. The splitting process was repeated five times to generate five independent training-validation-test partitions. Each partition had its own preprocessing pipeline to avoid information leakage.

Missing values were observed in wind direction and air pressure. These missing values were imputed using iterative imputation (Van Buuren and Groothuis-Oudshoorn 2011). An Extra Trees Regressor was used as the estimator in the round-robin imputation process. For each data split, the imputer was fitted only on the training set and then applied to the corresponding validation and test sets.

Continuous input features were standardized before being fed into the deep learning model. Standardization was applied to reduce the influence of scale differences across variables and to improve optimization stability. Binary regulation status indicators and categorical temporal variables were not standardized. To prevent information leakage, the standard scaler was fitted only on the training set for each split and then applied to the corresponding validation and test sets.

## 2.3. Bayesian Deep Learning Model Structure

The proposed Bayesian deep learning model integrates temporal meteorological and socio-economic covariates, calendar information, and regulation status indicators to predict daily $PM_{2.5}$ concentration while estimating regulation enforcement probability for propensity score adjustment. The model consists of five main Bayesian components: one Bayesian LSTM layer, two Bayesian embedding layers, and two Bayesian dense output layers. The model structure is defined in Equations (3)-(7).

$$\mathrm{h_t} = \text{Bayesian-LSTM}(\mathrm{x_t}, \mathrm{h_{t-1}}) \tag{3}$$

$$\mathrm{e_t^1} = \text{Bayesian-Embedding}(\mathrm{Month_t}, \text{Day of week}_\mathrm{t}) \tag{4}$$

$$\mathrm{e}_{\mathrm{t}}^{2} = \text{Bayesian-Embedding}(\mathrm{I}_{\mathrm{t}}) \tag{5}$$

$$\widehat{y_t} = \text{Bayesian-Dense}(\mathrm{h}_{\mathrm{t}}, \mathrm{e}_{\mathrm{t}}^{1}, \mathrm{e}_{\mathrm{t}}^{2}) \tag{6}$$

$$\widehat{I_t} = \text{Sigmoid}\Big(\text{Bayesian-Dense}(\mathrm{h}_{\mathrm{t}}, \mathrm{e}_{\mathrm{t}}^{1}, \mathrm{e}_{\mathrm{t}}^{2})\Big) \tag{7}$$

The first input stream contains meteorological and socioeconomic covariates. These covariates are fed into the Bayesian LSTM layer to capture temporal dependencies in the time-series data (Fortunato et al. 2017). The model uses an eight-day input window, including the previous seven days and the current day: $\mathrm{X}_{\mathrm{t}} = \{\mathrm{x}_{\mathrm{t}-7}, \ldots, \mathrm{x}_{\mathrm{t}}\}$. The second input stream contains two categorical temporal variables, including month and day of the week. These variables are mapped into a continuous temporal embedding $\mathrm{e}_{\mathrm{t}}^{1}$, allowing the model to account for seasonal patterns, weekly cycles, and other unobserved time-related effects (Yi et al. 2018). The third input stream contains the regulation-status vector: $I_t = \{I_t^1, \ldots, I_t^k\}$, where $k = 32$. Each element is a binary indicator, with 0 indicating that a regulation is not enforced on day $t$ and 1 indicating that it is enforced. This vector is transformed into a continuous regulation-status embedding $\mathrm{e}_{\mathrm{t}}^{2}$ using a separate Bayesian embedding layer (Pham et al. 2017).

The LSTM hidden representation $\mathrm{h}_{\mathrm{t}}$, temporal embedding $\mathrm{e}_{\mathrm{t}}^{1}$, and regulation status embedding $\mathrm{e}_{\mathrm{t}}^{2}$ are concatenated into a shared latent representation. This representation is then passed to two Bayesian dense output branches. The first branch predicts daily $PM_{2.5}$ concentration, denoted by $\widehat{y_t}$. The second branch predicts regulation enforcement probability, denoted by $\widehat{I_t}$, using a sigmoid activation function. This branch supports propensity score-based adjustment by estimating the probability of regulation enforcement conditional on observed covariates, thereby helping reduce confounding effects associated with non-random regulation implementation (Shi et al. 2019). A dropout layer is added to the RNN component to reduce overfitting. The dropout probability was set to 25% based on Cordova et al. (2021).

### 2.4. Bayesian Deep Learning Model Training

The proposed Bayesian deep learning model was trained to learn probability distributions over model parameters rather than fixed point estimates (Blundell et al. 2015). In each Bayesian layer, deterministic weights were replaced by posterior Gaussian distributions parameterized by a mean and standard deviation. A prior distribution was specified before training, and the variational posterior distribution was updated during training. This formulation allows uncertainty in the model parameters to be represented directly through the learned posterior weight distributions. In variational Bayesian learning, the learning objective is to approximate the true posterior distribution of the model weights $p(w|D)$ using a variational posterior distribution $q_\phi(w)$, where $D$ denotes the observed data and $\phi$ denotes the variational parameters. As shown in Equation (8), this objective is achieved by maximizing the evidence lower bound (ELBO):

$$\text{ELBO} = \mathrm{E}_{q_\phi(w)}[\log p(D|w)] - D_{KL}(q_\phi(w)||p(w)) \tag{8}$$

where $p(w)$ is the prior distribution of the weights, $p(D|w)$ is the likelihood of the observed data given the model weights, and $D_{KL}$ denotes the Kullback-Leibler divergence.

Maximizing the ELBO is equivalent to minimizing the negative ELBO: $\mathrm{E}_{q_\phi(w)}[-\log p(D|w)] + D_{KL}(q_\phi(w)||p(w))$. In the proposed model, the negative log-likelihood term contains two parts because the model has two outputs: daily $PM_{2.5}$ concentration and regulation status. The PM prediction output is treated as a continuous outcome under a Gaussian likelihood assumption, for which mean squared error (MSE) corresponds to the negative log-likelihood up to additive and multiplicative constants (Bishop and Nasrabadi 2006). The regulation status prediction output is treated as a binary outcome under a Bernoulli likelihood assumption, for which binary cross-entropy (BCE) corresponds to the negative log-likelihood (Terven et al. 2025). Therefore, as shown in Equation (9), the empirical training objective can be written as:

$$L = L_{MSE} + L_{BCE} + \frac{1}{N}\sum_{l=1}^{B} D_{KL}(q_{\varphi_l}(w_l)||p(w_l)) \quad (9)$$

where $N$ denotes the number of training observations, $B$ denotes the number of Bayesian layers, $q_{\varphi_l}(w_l)$ denotes the variational posterior distribution of the weights in layer $l$, and $p(w_l)$ denotes the prior distribution. The KL divergence loss was computed separately for each of the five Bayesian layers and scaled by $1/N$.

During each training epoch, Monte Carlo sampling was performed to sample the model parameters from their posterior distributions. The model outputs were then computed using the sampled parameters, and the MSE, BCE, and KL divergence losses were jointly optimized. Gradients were used to update the posterior means and standard deviations. For the two Bayesian embedding layers, backpropagation was performed using the reparameterization estimator (Kingma and Welling 2013). For the remaining Bayesian layers, the Flipout estimator was used to reduce gradient variance and improve training stability (Wen et al. 2018). Stochastic gradient descent was used for optimization. The number of training epochs was set to 30. The output dimension of the temporal embedding layer was set to 3 following (Yi et al. 2018), and the learning rate was set to 0.01. Hyperparameter tuning was conducted to select the final model configuration. The tuned hyperparameters included the output dimension of the regulation status embedding layer, the number of LSTM units, and the batch size. The candidate values were 3 and 5 for the regulation status embedding dimension, 128 and 256 for the number of LSTM units, and 32 and 64 for the batch size. These values produced eight combinations of hyperparameters. Each hyperparameter combination was evaluated ten times to account for stochastic variation during Bayesian training. In each run, the model was trained on the training set and evaluated on the validation set. The final configuration was selected based on the lowest mean validation MSE.

## 2.5. Regulatory Effect Estimation with Uncertainty Quantification

The regulatory effect was estimated by comparing observed $PM_{2.5}$ concentrations with counterfactual $PM_{2.5}$ concentrations under a hypothetical no-regulation scenario. In this study, the regulatory effect is defined as the average reduction in $PM_{2.5}$ attributable to air pollution regulations. Therefore, a positive estimated effect indicates that regulations reduced $PM_{2.5}$ concentration relative to the no-regulation counterfactual. To construct the no-regulation counterfactual, a regulation status vector containing only zeros was created: $I_0 = \{0, \dots, 0\}$. This vector has the same dimension as the original regulation status input and represents a hypothetical scenario in which none of the 32 regulations was enforced during the study period. The counterfactual prediction was generated by keeping the covariate input and temporal input unchanged while replacing the observed regulation status vector $I_t$ with the no-regulation vector $I_0$. This allows the model to estimate what daily $PM_{2.5}$ concentration would have been under the same meteorological, socioeconomic, and temporal conditions, but without regulation enforcement.

To account for uncertainties in Bayesian deep learning models, the prediction process was repeated across multiple trained posterior weight distributions (Kendall and Gal 2017). Specifically, the model was trained ten times for each of the five data splits, producing 50 fitted Bayesian models. Let $i \in [1,10]$ denotes the repeated training run within each split and $j \in [1,5]$ denotes the data split. For each fitted model $(i, j)$, the observed covariates and temporal variables were entered into the model, while the regulation status input was replaced by $I_0$. This produced a set of counterfactual PM predictions. The regulatory effect for each model run and data split was then calculated as the difference between the counterfactual $PM_{2.5}$ concentration and the observed $PM_{2.5}$ concentration:

$$ATE_{i,j} = \frac{1}{T}\sum_{t=1}^{T}[\hat{y}_t(I_0) - y_t] \quad (10)$$

where $y_t$ is the observed $PM_{2.5}$ concentration on day $t$, $\hat{y}_t(I_0)$ is the counterfactual prediction under the no-regulation scenario, and $T$ is the number of days in the evaluation period. Under this definition, a positive value of $ATE_{i,j}$ indicates that the predicted $PM_{2.5}$ concentration would have been higher without regulation, implying a reduction in $PM_{2.5}$ attributable to regulation.

To account for the non-Gaussian distribution of $PM_{2.5}$ concentrations and uncertainty across model runs, bootstrapping was applied to the estimated regulatory effects. The 50 estimated effects, $\{ATE_{1,1}, \dots, ATE_{10,5}\}$, were used as the empirical sampling distribution. In each bootstrap iteration, 50 ATE values were randomly resampled with replacement from this set, and the mean of the resampled ATE values was calculated. This procedure was repeated 10,000 times, producing a bootstrap distribution of mean regulatory effects. The final regulatory effect was summarized using the mean of the bootstrap distribution, and uncertainty intervals were obtained from the empirical percentiles of the bootstrapped estimates.

## 3. Results

### 3.1. Hyperparameter Selection and Performance Evaluation

Figure 4 presents the hyperparameter-tuning results based on validation MSE. Among the tested configurations, the combinations of (3, 128, 64) and (5, 128, 32) achieved comparable validation performance. Their mean validation MSE values were 62.812 and 62.565, respectively. Since the configuration (5, 128, 32) produced the lowest mean validation MSE, it was selected as the final hyperparameter setting for model evaluation.

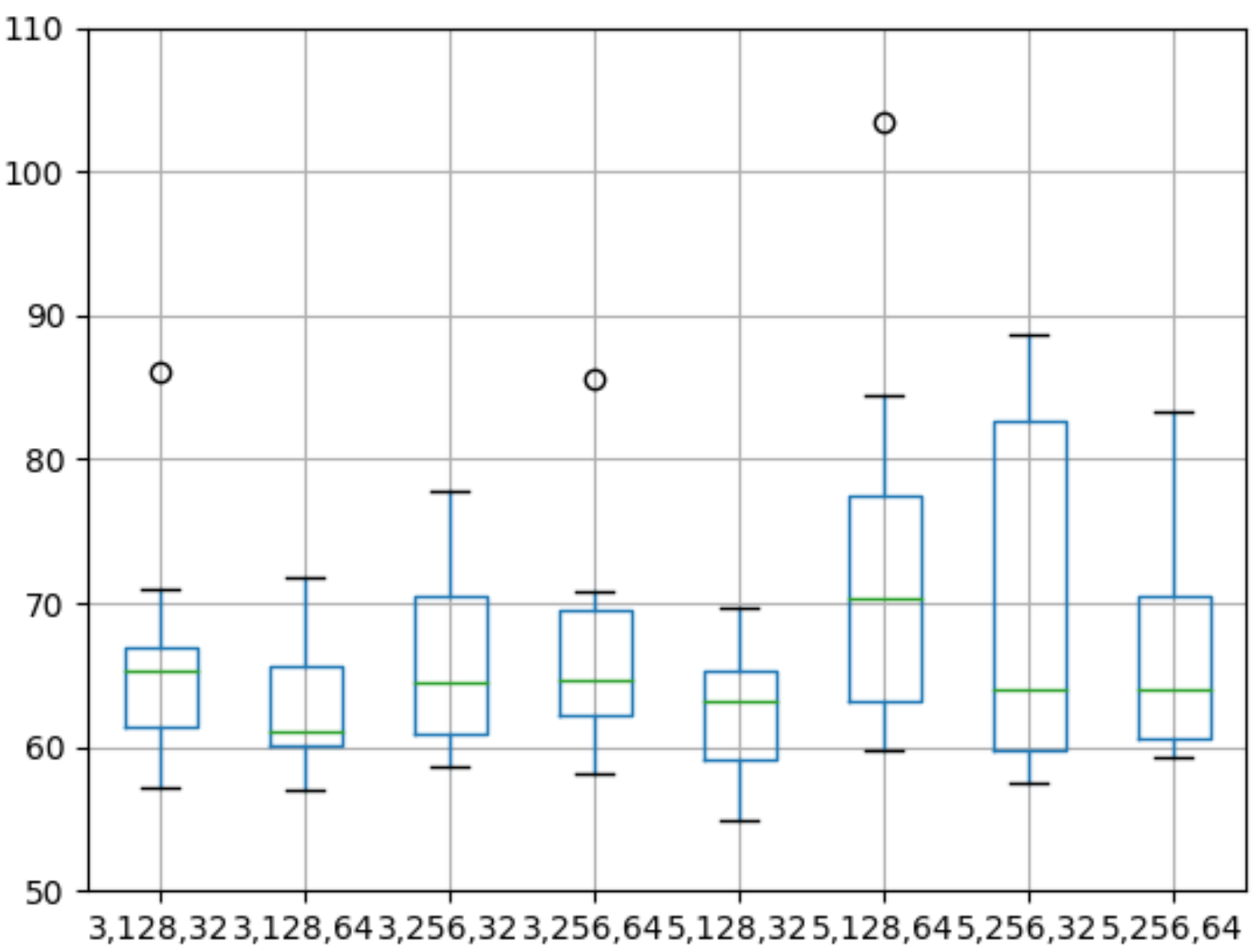

**Figure 4.** Hyperparameter tuning results based on validation mean squared error

Mean absolute error (MAE) and mean absolute percentage error (MAPE) were used to evaluate the $PM_{2.5}$ prediction performance of the proposed model. The evaluation was based on predicted $PM_{2.5}$ concentrations under the observed regulation status, rather than under the counterfactual no-regulation scenario. This ensured that the evaluation reflected the model's predictive accuracy for observed conditions.

Performance evaluation was conducted across all five data splits. For each split, the model was trained using the training set, while hyperparameters were selected based on validation MSE. The trained model was then used to predict $PM_{2.5}$ concentrations for the corresponding test set, and MAE and MAPE were calculated from the test-set predictions. Across the five data splits, the model achieved an average MAE of 4.52 and an average MAPE of 29.93%.

### 3.2. Regulatory Effect Analysis

Figure 5 compares the observed $PM_{2.5}$ concentrations with the counterfactual $PM_{2.5}$ concentrations estimated under a no-regulation scenario. Across the full study period, the mean counterfactual $PM_{2.5}$ concentration was 17.10 μg/m$^3$, with a 95% confidence interval of 16.86-17.34 μg/m$^3$. In comparison, the mean observed $PM_{2.5}$ concentration was 15.22 μg/$m^3$. This indicates an average regulatory effect of 1.88 μg/$m^3$, with a 95% confidence interval of 1.64-2.12 μg/$m^3$.

The difference between observed and counterfactual $PM_{2.5}$ concentrations was relatively small during the early years of the study period. From 2010 to 2013, the average difference was 0.95 μg/$m^3$, and in some months, the observed $PM_{2.5}$ concentration was higher than the counterfactual estimate. This suggests that the estimated regulatory effect was limited before 2013. After 2013, however, a clearer divergence emerged, with the counterfactual $PM_{2.5}$ concentration remaining consistently higher than the observed concentration. The gap widened further after 2018, indicating a stronger estimated regulatory effect in the later period. The largest monthly difference between counterfactual and observed $PM_{2.5}$ concentrations was 4.61 μg/$m^3$ in July 2018.

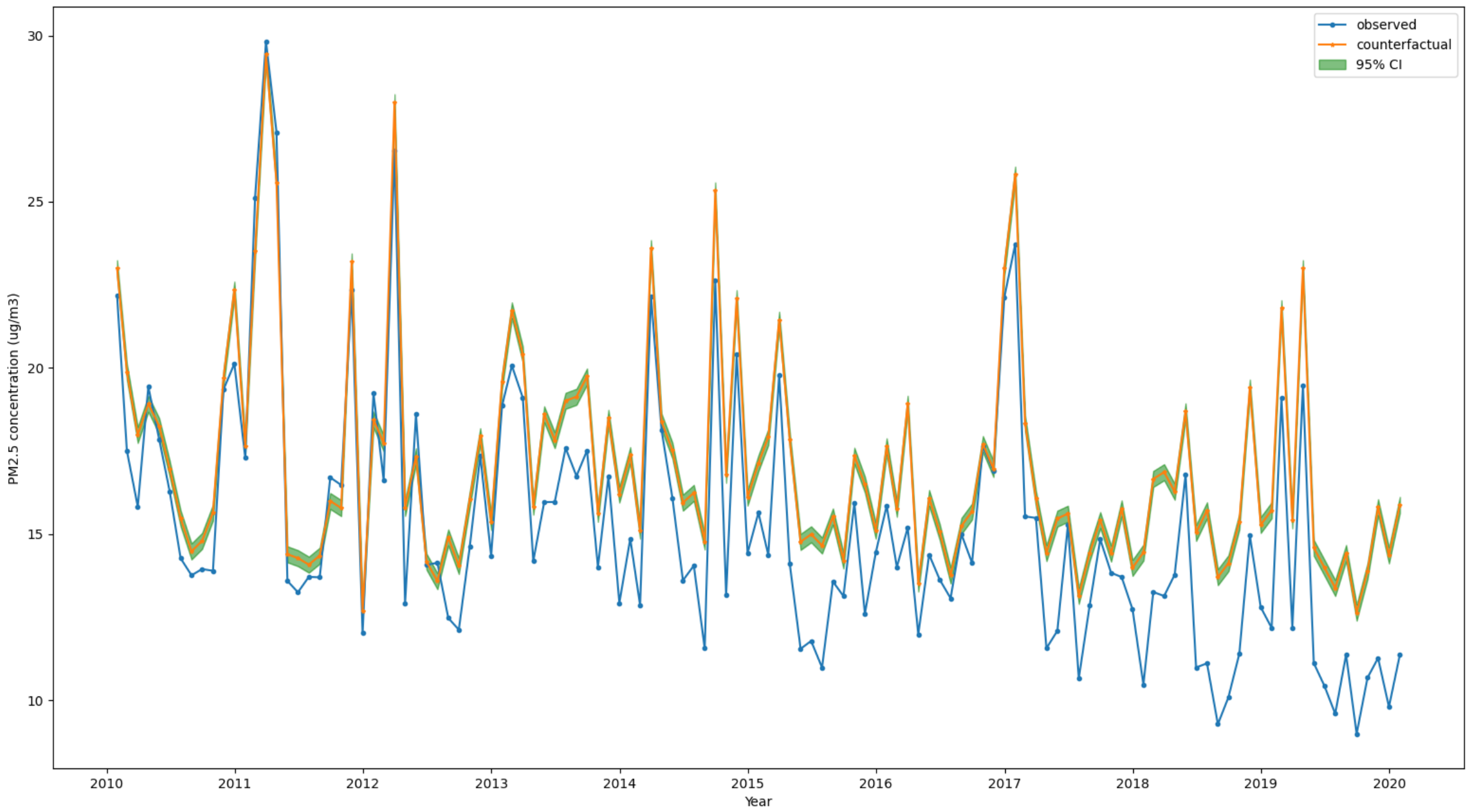

**Figure 5.** Comparison between observed $PM_{2.5}$ concentration and counterfactual $PM_{2.5}$ concentration under the no-regulation scenario during the study period

The estimated regulatory effects are further summarized in Table 2. The final counterfactual $PM_{2.5}$ estimates were calculated as the mean of 50 model outputs generated across the five data splits. The estimated $PM_{2.5}$ reduction remained modest before 2013. The annual reduction was slightly negative in 2011, at -0.02 μg/$m^3$, indicating that the observed annual $PM_{2.5}$ concentration was marginally higher than the estimated counterfactual concentration in that year. After 2013, the estimated reductions became consistently positive and generally larger. Over the full study period, air pollution regulations in London were associated with an average $PM_{2.5}$ reduction of 1.88 μg/$m^3$, corresponding to a relative reduction of 12.35%. The estimated effect varied across policy periods. Before 2013, the average reduction was relatively small, at 3.88%. From 2013 to 2017, the estimated regulatory effect increased to 12.46%, suggesting a stronger contribution of regulatory interventions to $PM_{2.5}$ reduction. The effect became substantially larger in 2018 and 2019, when the estimated relative reductions reached 29.42% and 29.32%, respectively. Taken together, these results suggest that air pollution regulations had a limited estimated effect in the early study period, but their contribution to $PM_{2.5}$ reduction became more pronounced after 2013 and especially after 2018.

**Table 2.** Comparison of observed and counterfactual $PM_{2.5}$ concentrations

| Year | Observed $PM_{2.5}$ (μg/$m^3$) | Counterfactual $PM_{2.5}$ (μg/$m^3$) | $PM_{2.5}$ Reduction (μg/$m^3$) | Relative $PM_{2.5}$ Reduction (%) |
|---|---|---|---|---|
| 2010 | 17.03 | 18.12 | 1.09 | 6.37 |

| 2011 | 18.43 | 18.41 | -0.02 | -0.09 |
|---|---|---|---|---|
| 2012 | 16.09 | 16.95 | 0.86 | 5.35 |
| 2013 | 16.64 | 18.52 | 1.88 | 11.28 |
| 2014 | 16.16 | 18.27 | 2.11 | 13.07 |
| 2015 | 13.99 | 16.46 | 2.46 | 17.60 |
| 2016 | 15.33 | 16.61 | 1.29 | 8.40 |
| 2017 | 14.36 | 16.08 | 1.72 | 11.96 |
| 2018 | 12.34 | 15.96 | 3.63 | 29.42 |
| 2019 | 12.18 | 15.75 | 3.57 | 29.32 |
| All | 15.22 | 17.10 | 1.88 | 12.35 |

## 4. Discussion

### 4.1. Key Findings

This study developed an uncertainty-aware Bayesian deep learning framework to estimate the regulatory effect of air pollution policies on $PM_{2.5}$ concentrations in London from 2010 to January 2020. By comparing observed $PM_{2.5}$ concentrations with counterfactual estimates under a hypothetical no-regulation scenario, the model provides evidence that air pollution regulations contributed to measurable reductions in $PM_{2.5}$ concentrations over the study period.

The results show that the average observed $PM_{2.5}$ concentration was 15.22 μg/m$^3$, while the average counterfactual $PM_{2.5}$ concentration was estimated to be 17.10 μg/m$^3$. This corresponds to an average regulatory effect of 1.88 μg/m$^3$, with a 95% confidence interval of 1.64-2.12 μg/m$^3$, and an overall relative $PM_{2.5}$ reduction of 12.35%. These findings suggest that, in the absence of air pollution regulations, $PM_{2.5}$ concentrations in London would likely have remained higher during the study period.

The estimated regulatory effect was not uniform over time (see Figure 5). Before 2013, the difference between observed and counterfactual $PM_{2.5}$ concentrations was relatively small, indicating that the estimated effect of regulations was limited in the early study period. From 2013 to 2017, the estimated effect became more evident, with an average relative $PM_{2.5}$ reduction of 12.46%. The strongest estimated effects occurred in 2018 and 2019, when the relative reductions reached approximately 29%. This temporal pattern suggests that regulatory effects may accumulate over time and become more pronounced when policy measures are strengthened, expanded, or implemented in combination.

The yearly results (see Table 2) also show some variation in estimated effects. For example, the estimated reduction was slightly negative in 2011, indicating that observed $PM_{2.5}$ was marginally higher than the counterfactual estimate in that year. This finding highlights the importance of evaluating policy effects over multiple years rather than relying on single-year comparisons. Air pollution concentrations are influenced by meteorology, socioeconomic activity, transport patterns, and other time-varying factors. Therefore, short-term deviations do not necessarily indicate regulatory failure, but they do demonstrate the need for models that can account for temporal complexity and uncertainty.

Several limitations should be acknowledged. First, this study estimated aggregate regulatory effects rather than the independent effect of each individual policy, because multiple regulations overlapped during the study period. Second, although the model includes meteorological, socioeconomic, and regulation status variables, unobserved confounders may still affect the estimated counterfactual outcomes. Third, $PM_{2.5}$ observations were aggregated across selected Inner London monitoring stations, which may obscure local spatial heterogeneity. Future research could extend the framework to estimate policy-specific and spatially differentiated effects, incorporate additional transport and emissions data, and compare results across multiple cities.

### 4.2. Policy Implications

The findings have several implications for air pollution governance and regulatory evaluation. First, the results suggest that sustained air pollution regulation can produce measurable public environmental benefits. The increasing estimated effect after 2013, and especially after 2018, indicates that regulatory interventions may require time to generate substantial effects. This supports the need for long-term policy commitment rather than short-term or fragmented interventions.

Second, the results imply that policy effectiveness is likely strengthened when regulations are implemented as part of a broader and cumulative policy package. Air quality improvement rarely results from a single intervention. Instead, reductions in $PM_{2.5}$ may reflect the combined effects of traffic control, emissions regulation, cleaner transport measures, industrial and commercial emissions management, and broader urban environmental planning. The stronger estimated reductions in the later years suggest that integrated and reinforced regulatory strategies may be more effective than isolated measures.

Third, the temporal heterogeneity in regulatory effects highlights the importance of continuous policy monitoring. The limited estimated effect before 2013 and the stronger effect after 2018 suggest that policy evaluation should not only ask whether regulation works, but also when, how strongly, and under what conditions it works. This is particularly important for urban air quality governance, where policy implementation, behavioral adaptation, technological change, and environmental conditions may interact over time.

Fourth, the counterfactual results can support more evidence-based policy design. By estimating what $PM_{2.5}$ concentrations might have been under a no-regulation scenario, the model provides a more policy-relevant measure than descriptive trend analysis alone. Such estimates can help policymakers assess the magnitude of regulatory benefits, identify periods of stronger or weaker policy performance, and adjust future interventions accordingly.

Finally, the results suggest that uncertainty-aware methods are especially important for environmental policy evaluation. Since regulatory exposure is not randomly assigned and air pollution is shaped by multiple confounding factors, policy evaluation should explicitly account for uncertainty. The Bayesian framework used in this study provides not only point estimates of regulatory effects but also confidence intervals, allowing decision-makers to interpret results with appropriate caution.

### 4.3. AI for Social Good Significance

This study contributes to AI for Social Good by applying causal AI to an environmental policy problem with direct implications for public health, regulatory accountability, and urban sustainability. Rather than using AI only to predict pollution levels, this study uses an uncertainty-aware Bayesian deep learning framework to estimate the counterfactual effect of air pollution regulations. This shifts the role of AI from environmental forecasting to policy-relevant decision support.

This work builds on a sustained research trajectory on AI-enabled environmental and health decision-making. Earlier studies established the broader AI for Social Good agenda by examining how AI and big data can support environmental decision-making, personalized air pollution monitoring, and health management (Li et al. 2021a; Li et al. 2021b). Subsequent work developed the data and modeling foundations needed for this agenda, including citizen-centric sensor placement for air quality monitoring, image-based real-time air pollution estimation, domain-specific Bayesian air pollution forecasting, fine-grained urban air pollution estimation and forecasting, missing air pollution data recovery, universal air pollutant sensor calibration, and image-based probabilistic ambient air pollution estimation under limited-data conditions (Han et al. 2020; Han et al. 2025; Han et al. 2024b; Song et al. 2020; Sun et al. 2019; Yu et al. 2025; Zhang et al. 2022).

A parallel stream of this research has connected environmental AI with public health and causal inference. Prior studies examined the relationship between outdoor $PM_{2.5}$ exposure and COVID-19 infection dynamics and used interpretable AI-driven causal inference to estimate time-varying effects of $PM_{2.5}$ and public health interventions on COVID-19 infection rates (Han et al. 2024a; Han et al. 2021b). Most directly, Han et al. (2021a) developed a Bayesian LSTM regulatory intervention framework to estimate the effects of air pollution control regulations in Beijing. The present London study extends this line of work to a new policy and geographic context, demonstrating how uncertainty-aware AI can be used not only for prediction, but also for counterfactual environmental policy evaluation.

From an AI for Social Good perspective, the value of the proposed framework lies in its ability to make air pollution policy evaluation more transparent, evidence-based, and decision-relevant. By estimating what $PM_{2.5}$ concentrations might have been under a no-regulation scenario, the model provides a policy-relevant measure that goes beyond descriptive trend analysis. This can help policymakers and the public assess whether observed air quality improvements are plausibly attributable to regulatory action, identify periods of stronger or weaker policy effectiveness, and inform future environmental governance. More broadly, this study shows how AI can support socially beneficial decision-making when predictive modeling is combined with causal reasoning, uncertainty representation, and public policy relevance.

## 5. Conclusion

This study developed an uncertainty-aware Bayesian deep learning framework to estimate the aggregate effect of air pollution regulations on PM2.5 concentrations in London from 1 January 2010 to 31 January 2020. By integrating daily air quality observations, meteorological covariates, socioeconomic indicators, month-of-year and day-of-week indicators, and regulation status data, the proposed model estimated counterfactual $PM_{2.5}$ concentrations under a hypothetical no-regulation scenario. The results show that London's air pollution regulations were associated with an average $PM_{2.5}$ reduction of 1.88 $\mu g/m^3$, corresponding to a relative reduction of 12.35%. The estimated regulatory effect was limited in the early study period, became more evident after 2013, and was strongest in 2018 and 2019. These findings suggest that sustained and cumulative regulatory interventions can contribute to measurable improvements in urban air quality, particularly when policies are strengthened, expanded, and implemented over time. Methodologically, this study extends the Bayesian LSTM regulatory intervention framework from Beijing to London, demonstrating its applicability in a distinct urban governance and policy context. By combining counterfactual prediction, propensity score-based adjustment, and Bayesian uncertainty representation, the framework provides a policy-relevant approach for evaluating environmental regulations using observational data. More broadly, this study contributes to AI for Social Good by showing how uncertainty-aware causal AI can support public health protection, environmental accountability, and evidence-based urban governance. Rather than using AI only for air pollution forecasting, the proposed approach uses AI to estimate whether regulations changed pollution trajectories and how uncertain those estimates are. Overall, this study demonstrates the potential of uncertainty-aware causal AI as a decision-support tool for evaluating whether urban environmental regulations deliver measurable social and public health benefits.

**Declaration of generative AI use.** The authors acknowledge the use of generative AI tools to support the writing process. ChatGPT-5.5 was used to improve the language of the manuscript and to assist in the revision of the manuscript. The authors reviewed the content generated and took full responsibility for the content of the submitted manuscript.

**Acknowledgments.** The authors gratefully acknowledge London Air, Meteostat, the U.S. National Oceanic and Atmospheric Administration, the UK Department for Transport, the Office for National Statistics, the Department for Environment, Food & Rural Affairs, and The National Archives for making the data used in this study publicly available.

**Funding statement.** This research is supported in part by the General Research Fund (GRF) of the Research Grants Council of Hong Kong, under Grant No. 17620920. The funder had no role in study design, data collection and analysis, decision to publish, or preparation of the manuscript.

**Competing interests.** None.

**Data availability statement.** All data used in this study were obtained from publicly accessible sources. Detailed descriptions of data sources are provided in Section 2.1.

**Author contributions.** Conceptualization: JCK Lam; VOK Li. Methodology: YW Man; JCK Lam; VOK Li; Y Han. Data curation: YW Man. Visualization: YW Man. Writing original draft: YW Man; JCK Lam; VOK Li; Y Han. Funding acquisition: JCK Lam; VOK Li. All authors approved the final submitted draft.